\documentclass[conference,a4paper]{IEEEtran}

\usepackage{amsmath,amssymb,amsfonts}
\usepackage{balance}
\usepackage{bm}
\usepackage{booktabs}
\usepackage{cite}
\usepackage{enumitem}
\usepackage[T1]{fontenc}
\usepackage{graphicx}
\usepackage{multirow}
\usepackage{siunitx}
\DeclareSIUnit{\belmilliwatt}{Bm}
\DeclareSIUnit{\dBm}{\deci\belmilliwatt}
\usepackage{soul}
\usepackage{textcomp}
\usepackage{threeparttable}
\usepackage[color=green!25]{todonotes}
\soulregister\cite7
\soulregister\footnote7
\soulregister\eqref7
\soulregister\ref7
\usepackage{xcolor}
\usepackage{url}
\usepackage{hyperref}
\graphicspath{{./}{../figure/}}

\begin{document}

\title{%
  Static vs. Dynamic Databases for Indoor Localization based on Wi-Fi
  Fingerprinting:\\
  A Discussion from a Data Perspective}%

\author{%
  \IEEEauthorblockN{Zhe Tang\IEEEauthorrefmark{1}\IEEEauthorrefmark{2}, Ruocheng
    Gu\IEEEauthorrefmark{1}, Sihao Li\IEEEauthorrefmark{1}\IEEEauthorrefmark{2},
    Kyeong Soo Kim\IEEEauthorrefmark{1}, and Jeremy S.
    Smith\IEEEauthorrefmark{2}}%
  \IEEEauthorblockA{\IEEEauthorrefmark{1}School of Advanced Technology,
    Xi'an Jiaotong-Liverpool University, Suzhou 215123, China.\\
    Email: \{Zhe.Tang15,~Ruocheng.Gu22,~Sihao.Li19\}@student.xjtlu.edu.cn,
    Kyeongsoo.Kim@xjtlu.edu.cn}
  \IEEEauthorblockA{\IEEEauthorrefmark{2}Department of Electrical Engineering
    and Electronics,
    University of Liverpool, Liverpool, L69 3GJ, U.K.\\
    Email: \{Zhe.Tang,~Sihao.Li,~J.S.Smith\}@liverpool.ac.uk}%
}%

\maketitle

\begin{abstract}
  Wi-Fi fingerprinting has emerged as the most popular approach to indoor
  localization because it does not require deployment of new infrastructure or
  the modification of existing systems but exploits Wi-Fi networks already
  deployed in most indoor environments. The use of machine learning algorithms,
  including deep neural networks (DNNs), has greatly improved the localization
  performance of Wi-Fi fingerprinting, but its success heavily depends on the
  availability of fingerprint databases composed of a large number of the
  received signal strength indicators (RSSIs) measured at reference points, the
  medium access control addresses of access points, and the other available
  measurement information. However, most fingerprint databases do not reflect
  well the time varying nature of electromagnetic interferences in the more
  complicated modern indoor environment due to the increase in Wi-Fi and
  Bluetooth equipment. This could result in significant changes in statistical
  characteristics of training/validation and testing datasets, which are often
  constructed at different times, and even the characteristics of the testing
  datasets could be different from those of the data submitted by users during
  the operation of localization systems after their deployment. In this paper,
  we consider the implications of time-varying Wi-Fi fingerprints on indoor
  localization from a data-centric point of view and discuss the differences
  between static and dynamic databases. As a case study, we have constructed a
  dynamic database covering three floors of the International Research building
  on the south campus of Xi'an Jiaotong-Liverpool University (XJTLU) based on
  RSSI measurements, over 44 days, and investigated the differences between
  static and dynamic databases in terms of statistical characteristics and
  localization performance. The analyses based on variance calculations and
  Isolation Forest show the temporal shifts in Wi-Fi RSSIs, which result in a
  noticeable trend of the increase in the localization error of a Gaussian
  process regression model with the maximum error of 6.65~m after 14 days of
  training without model adjustments. The results of the case study with the
  XJTLU dynamic database clearly demonstrate the limitations of static databases
  and the importance of the creation and adoption of dynamic databases for
  future indoor localization research and real-world deployment.
\end{abstract}


\section{Introduction}
\label{sec:introduction}
%%%
Wi-Fi fingerprinting has become the most popular technology for indoor
localization, which can leverage existing Wi-Fi network infrastructure to
provide reliable indoor localization services. The use of machine learning
algorithms---especially deep neural networks (DNNs)---has brought substantial
improvement in the performance and scalability of indoor localization based on
Wi-Fi fingerprinting~\cite{Robustness,DNN}.

Complicated and time-varying indoor electromagnetic interferences, however, pose
a serious challenge to the robustness of indoor localization models. As indoor
electromagnetic interferences are highly time-varying~\cite{multi-sensor}, it is
desirable to develop more robust indoor localization models based on databases
that can take into account the time variability of Wi-Fi fingerprints over days,
weeks, or even longer. Nevertheless, due to the higher labor cost of
constructing such Wi-Fi fingerprint databases, researchers often end up with
validating their algorithms based on closed, private databases typically
covering a single floor with a smaller number of access points (APs) and
reference points (RP) and neglecting temporal signal fluctuations.

A better alternative is to use well-known, publicly-available databases like
those summarized in Table~\ref{tbl:public-dbs}, which enables fair comparison
among indoor localization algorithms. In Table~\ref{tbl:public-dbs}, we classify
the databases into two groups, i.e., dynamic and static databases. Dynamic
databases are defined by their incorporation of temporal variations in received
signal strengths (RSSs) or received signal strength indicators
(RSSIs)\footnote{As a relative indicator, RSSI has no unit unlike RSS whose
  typical unit is \si{dBm}.} over a long period of time. Such databases provide
the periodic measurements of RSSs/RSSIs over a predefined set of RPs to
ascertain the continual relevance and timeliness of the recorded data. Static
databases, on the other hand, are characterized by the absence of such periodic
measurements. In this regard, databases with training and test datasets measured
at different times (e.g., UJIIndoorLoc~\cite{UJI}) are considered as static
databases. Note that the existence of auxiliary information on APs, such as
channel specifications, are not considered in this classification.
%%%
\begin{table*}[]
\centering
\caption{Publicly-available Wi-Fi fingerprint databases.}
\label{tbl:public-dbs}
\begin{tabular}{cccccc}
\hline
Database                                                    & No. of APs & Covered Area [\si{\meter\squared}] & Category & Frequency Band & Year \\ \hline
UJIIndoorLoc~\cite{UJI}                                     & 520        & 108703         & Static   & \SI{2.4}{\GHz}        & 2014 \\
UJI LIB DB~\cite{LongTermWiFiFingerprinting}                & 448        & N/A            & Dynamic  & \SI{2.4}{\GHz}        & 2018 \\
Tampere University~\cite{data2040032}                       & 992        & 22570          & Dynamic  & 2.4/5~\si{\GHz}      & 2017 \\
WI-FI RSSI Indoor Localization~\cite{1yd5-rn96-19}          & 6          & 384            & Static   & 2.4/5~\si{\GHz}      & 2019 \\
WI-FI Fingerprinting Radio Map Database~\cite{sdpr-fh91-23} & 695        & 717            & Static   & N/A            & 2023 \\
Hybrid Dataset~\cite{data7110156}                           & 17         & 386            & Static   & \SI{2.4}{\GHz}        & 2022 \\
MTLoc~\cite{MTLoc}                                          & 3365       & 8350           & Dynamic  & 2.4/5~\si{\GHz}      & 2023 \\ \hline
\end{tabular}
\end{table*}
%%%

In this paper, we consider the implications of time-varying Wi-Fi fingerprints
on indoor localization from a data-centric point of view and discuss the
limitations of static databases for indoor localization based on a careful
review of the existing publicly-available databases. We also present the results
of systematic investigation of a dynamic database covering three floors of the
International Research (IR) building on the south campus of Xi’an
Jiaotong-Liverpool University (XJTLU) composed of RSSI measurements over 44
days. Our work, in this paper, highlights the importance of the creation and
adoption of dynamic databases for indoor localization and provides researchers
valuable guidelines for the construction of dynamic databases.

\section{Issues in the Construction of Static Databases}
\label{sec:static-db-issues}
%%%
In practice, static Wi-Fi fingerprint databases are constructed in such a way
that the RSSI measurement times of their training, validation, and testing
datasets are different from one another. The different RSSI measurement times
may result in shifts in the statistical characteristics of their RSSI
fingerprints due to changes in APs, time-varying wireless channels, and
imperfect measurement practice, which could worsen the actual localization
performance with the testing dataset of a model trained and fine-tuned with the
training and validation datasets.

\subsection{Changes in APs}
\label{sec:aps}
%%%
The major causes for the changes in APs are as follows:
%%%
\begin{itemize}
\item \textbf{Mobile hotspots}: Nowadays, many mobile devices can work as Wi-Fi
  hotspots, which causes significant issues in Wi-Fi fingerprinting based on
  static databases. Due to their sporadic operations and moving with users, the
  detection of mobile hotspots highly depends on measurement time and
  location. Moreover, some hotspots can employ ephemeral MAC addresses upon each
  initiation, so the resulting RSSIs could be regarded as those from multiple,
  different APs. Proper handling of these mobile hotspots (e.g., filtering them
  out in the datasets), however, is quite challenging with conventional static
  databases.
\item \textbf{Addition and failure of fixed APs}: As typical APs are deployed
  and installed in fixed and often difficult-to-access locations of a building
  (e.g., ceiling), the measurement of their RSSIs is confined to the
  neighborhood of their deployment. Still, APs could be replaced by new ones due
  to device malfunction or as part of network upgrade. Also, new APs could be
  deployed in addition to the existing ones. Unlike mobile hotspots, however,
  the changes in fixed APs are rare and hardly captured in a short time frame.
\item \textbf{Network Maintenance}: In large-scale building complexes like
  shopping malls and office buildings, network maintenance may be done on a
  periodical basis to check and improve the functionality of the whole network
  infrastructure. During the maintenance, a network operator can change the
  attributes (e.g., service set identifiers (SSIDs) and channels) and the
  operation mode (e.g., active, standby, and sleep) of managed APs, which,
  again, would result in differences in the statistical characteristics of the
  RSSI fingerprints of the datasets.
\end{itemize}
%%%

\subsection{Time-Varying Wireless Channels}
\label{sec:tv-wc}
%%%
In addition to the changes in APs, time-varying wireless channels affect RSSIs
as well, which could result from the following phenomena:
%%%
\begin{itemize}
\item \textbf{Multipath Propagation}: It is likely that in an indoor
  environment, radio signals from an AP reaches the receiving antenna of a user
  device through multiple paths due to reflections on the surfaces of objects
  like furniture, walls, and ceilings. This multipath propagation causes
  multipath interference, which is a highly time-varying phenomenon and thereby
  results in time-varying wireless channels.
\item \textbf{Dynamic Disturbance}: Another major cause of time-varying wireless
  channels is dynamic disturbance from devices like Bluetooth devices, microwave
  ovens, and wireless microphones operating in the same frequency bands as Wi-Fi
  APs, which negatively affects Wi-Fi RSSIs at receivers through co-channel
  interference.
\item \textbf{Environmental Changes}: Environmental changes like those in
  atmospheric temperature and humidity result in the dispersion, diffraction,
  and absorption of electromagnetic signals, while lightning could cause
  electromagnetic pulse disruptions. Their combined effects make wireless
  channels time-varying, too.
\end{itemize}
%%%

\subsection{Imperfect Measurement Practice}
\label{sec:measurement-practice}
%%%
The measurement practice, too, could adversely affect the consistency in the
statistical characteristics of RSSI fingerprints unless carefully planned and
applied.
%%%
\begin{itemize}
\item \textbf{Measurement Devices}: The number of \SI{5}{\GHz}-enabled Wi-Fi
  devices has been continuously growing. For dual-band APs, a singular MAC
  address is used for both \SI{2.4}{\GHz} and \SI{5}{\GHz} bands. Most of the
  fingerprint databases constructed before the widespread adoption of
  \SI{5}{\GHz} devices, by the way, do not provide channel information. Also,
  the computation of RSSI values is not consistent among equipment
  manufacturers, most of which do not provide explicit information on
  it. Additionally, the computation of RSSI values on the same device could be
  different depending on the versions of firmware and operating systems.
\item \textbf{Measurement RPs}: When the same RPs (e.g., those based on a fixed
  grid on a floor) are used for the construction of the datasets over a period,
  measuring RSSIs at the same RPs at different times could also make the
  acquired RSSIs time-varying (e.g., due to different poses and directions),
  which further exacerbates the issues of static databases already discussed.
\end{itemize}
%%%

\section{A Case Study: XJTLU Dynamic Database}
\label{sec:xjtlu-dynamic-db}
%%%
Having discussed the issues in the construction of conventional static
fingerprint databases in Section~\ref{sec:static-db-issues}, here we demonstrate
how a dynamic database can be utilized to systematically investigate the temporal
aspects of Wi-Fi RSSI fingerprints and their impact on indoor localization
performance through a case study based on the XJTLU dynamic database.

\subsection{Experimental Setup}
\label{sec:exp-setup}
%%%
For this case study, we constructed a new dynamic Wi-Fi RSSI fingerprint
database covering three floors of the IR building on the south campus of XJTLU.

Building a dynamic database requires repeated access to the same RPs over a long
period of time (e.g., one month), so we adopted a hybrid measurement scheme;
users carrying laptops and Android smartphones visited the assigned RPs and
measured RSSIs on a daily basis, while Raspberry~Pi~Pico~Ws mounted on the
corridor walls measured RSSIs automatically every hour. Although the
Raspberry~Pi~Pico~W supports only the \SI{2.4}{\GHz}-band Wi-Fi, we selected it as
an anchor device due to the right balance between power consumption and
performance. Table~\ref{tbl:database description} provides an overview of the
XJTLU single-building multi-floor dynamic fingerprint database.

\begin{figure}[!tbh]
  \centering%
  \includegraphics[width=\linewidth]{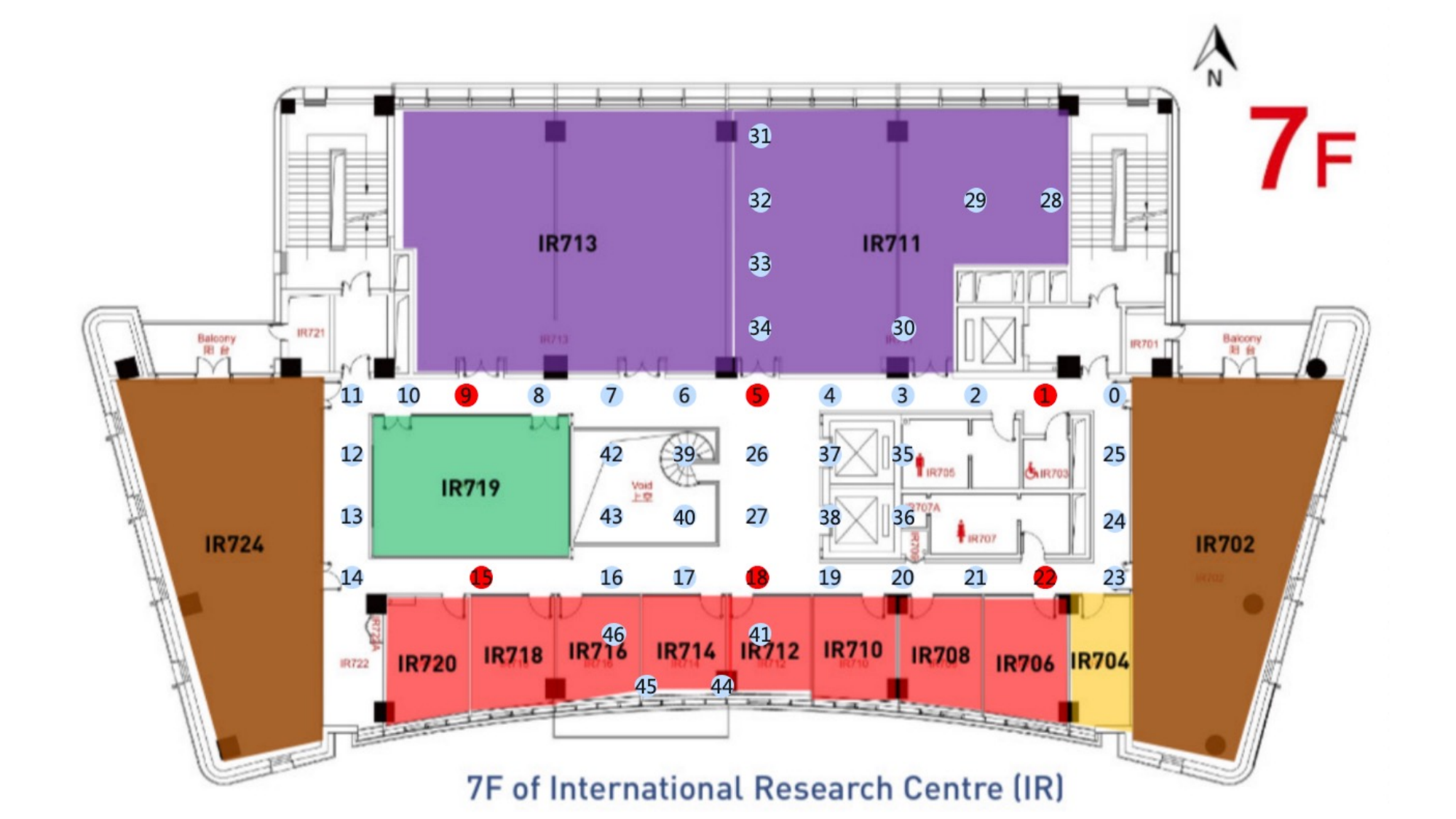}
  \caption{RP distribution on the 7th floor of the IR building; the RPs with
    Raspberry~Pi~Pico~W are marked in red.}
  \label{fig:RP-map}
\end{figure}

The distribution of RPs is shown in Fig~\ref{fig:RP-map}, where the red markers
indicate the RPs for Raspberry~Pi~Pico~Ws. The spacing between RPs, which is
about \SI{3}{\m}, is not strict because we put each RP in close proximity to a
landmark of the building to ease the task of repetitive RSSI measurements
without compromising the normal use of the building. The coordinates of an RP
are relative to the reference point of RP~No.~0 and given in meters. The number
of RPs per floor is also summarized in Table~\ref{tbl:database description}. Due
to the differences in the floor structures, the numbers of APs for the three
floors are different from one another. The average width of the building's
corridors is around \SI{1.3}{\m}, and their walls are made of glass. The walls
between the rooms, on the other hand, are of brick construction. In the centers
of the 6th and the 7th floor, there is a corkscrew staircase connecting the two,
while the center of the 8th floor is an open plan space.

%%%
\begin{table*}[]
  \begin{center}
    \caption{Overview of the XJTLU dynamic database.}
    \label{tbl:database description}
    \begin{threeparttable}
      \begin{tabular}{cccclcccc}
        \hline
        Building            & Floor & RP Numbers  & \multicolumn{2}{c}{Devices}                                                                               & Period /day                   & AP                   & Samples                 & Covered Area [\si{\meter\squared}]             \\ \hline
        \multirow{3}{*}{IR} & 6     & 0--27        & 6 Raspberry Pi Pico W & \multirow{3}{*}{\begin{tabular}[c]{@{}l@{}}1 laptop/1 smartphone\tnote{*}\end{tabular}} & \multirow{3}{*}{44 } & \multirow{3}{*}{446} & \multirow{3}{*}{511237} & \multirow{3}{*}{1200} \\
                            & 7     & 0--34        & 6 Raspberry Pi Pico W &                                                                                  &                          &                      &                         &                    \\
                            & 8     & 0--25, 35--46 &                       &                                                                                  &                          &                      &                         &                    \\ \hline
      \end{tabular}
      \begin{tablenotes}
      \item[*] The laptop and the smartphone are MacBook Pro and Xiaomi 13,
        respectively, and were used for all the floors.
      \end{tablenotes}
    \end{threeparttable}
  \end{center}
\end{table*}
%%%

%%% 
%%%

\subsection{Statistical Characteristics}
\label{sec:statistics}
%%%
We undertook statistical analyses of the time slices of XJTLU dynamic database
based on a machine learning algorithm called Isolation
Forest~\cite{liuIsolationForest2008} as well as conventional techniques to
investigate temporal aspects of RSSI fingerprints.

\subsubsection{RSSI Variation over Time}
\label{sec:simple-time-dimension-statistics}
\begin{figure}[!tbh]
  \centering%
  \includegraphics[width=.9\linewidth,trim=170 45 150 55, clip]{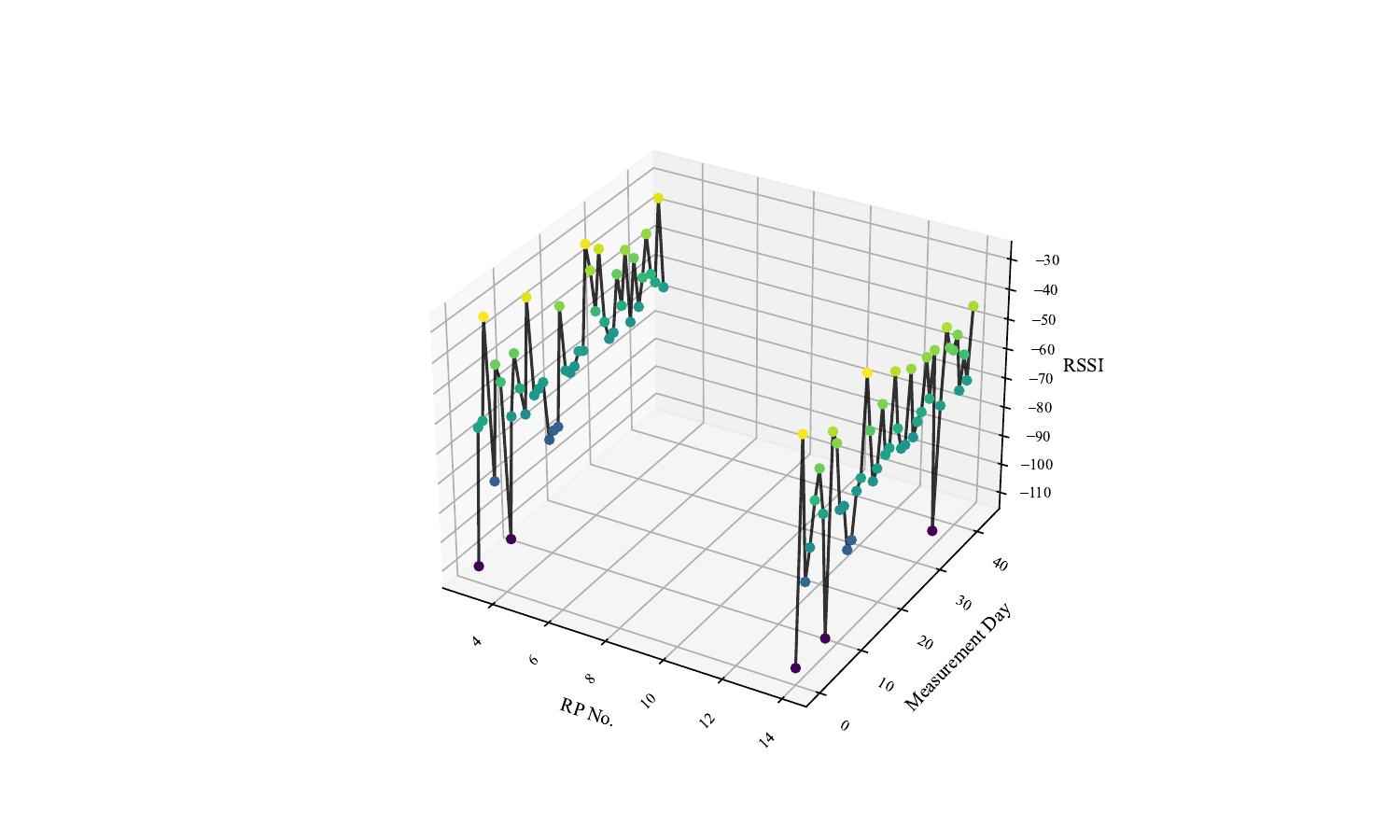}
  \caption{Time variation of the RSSIs from AP~No.~79 measured at RPs No.~3 and
    14 on the 7th floor of the IR building.}
  \label{fig:RSSI_time}
\end{figure}

%%%
Fig~\ref{fig:RSSI_time} shows the time variation of the RSSIs from AP~No.~79
measured at RPs~No.~3 and 14 on the 7th floor of the IR building. This
combination of an AP and RPs is taken as an example because the range of the
RSSI values for this case---i.e., $[-110,-25]$\footnote{As in \cite{DNN}, we use
  -110 as an RSSI value to indicate no detection of an AP.}---is the largest.

Note that, though RPs~No.~3 and 14 are located far from each other and thereby
have different wireless channel characteristics, the two time series of RSSIs
show similar patterns and that their statistical variances of 324.89 and 324.91
at RPs~No.~3 and 14 are also close to each other. These imply that there is a
potential correlation among RSSIs from the same AP over time, which could be
exploited to improve the performance of indoor localization based on dynamic
data bases.
%%% 
%%%

\subsubsection{RSSI Anomaly Detection based on Isolation Forest}
\label{sec:isolation-forest}
%%%
We carried out RSSI time series anomaly detection based on the Isolation Forest
algorithm~\cite{liuIsolationForest2008} to demonstrate that, because RSSIs could
be significantly abnormal at certain points of time, the mean RSSI value cannot
properly represent the whole RSSI time series. The core idea is that normal
samples can be easily separated by the decision tree of the Isolation Forest,
while abnormal samples are more difficult to separate, and thus, their path
lengths and isolation degrees become larger. In the Isolation Forest, an anomaly
score $s(x,n){\in}[0,1]$ indicates the degree of anomaly in the sample, which is
defined as follows:
%%%
\begin{equation}
  \label{eq:anomaly score}
  s(x,n) = 2^{-\frac{\operatorname{E}[h(x)]}{c(n)}}, 
\end{equation}
%%%
where $h(x)$ denotes the length of the path from the root node to the leaf node
containing a sample $x$, and $c(n)$ is a function of the number of samples $n$;
if $s$ is close to 1, then the sample $x$ is very likely to be an
outlier~\cite{liuIsolationForest2008}.

We used the implementation of the Isolation Forest algorithm provided by the
scikit-learn Python package with the parameter settings summarized in
Table~\ref{tbl:Isolation Forest parameters setting}.
%%%
\begin{table}[!tb]
\centering
\caption{Hyperparameters and their values for Isolation Forest.}
\label{tbl:Isolation Forest parameters setting}
\begin{tabular}{ccc}
\hline
Parameter Settings & Value     & Description                                                                                 \\ \hline
n\_estimators      & 100       & number of random trees                                                                      \\
contamination      & 0.10 & \begin{tabular}[c]{@{}c@{}}percentage of \\ anomalous data\end{tabular}                     \\
max\_samples       & auto      & \begin{tabular}[c]{@{}c@{}}number of samples to \\ construct the subtree\end{tabular}       \\
max\_features      & 1.0       & \begin{tabular}[c]{@{}c@{}}constructing number \\ of features for each subtree\end{tabular} \\
random\_state      & 42        & random seed                                                                                 \\ \hline
\end{tabular}
\end{table}
%%%
Note that, unlike the anomaly score defined in \eqref{eq:anomaly score}, the
anomaly score returned from the scikit-learn implementation provides a negative
value for outliers~\cite{IsolationForest}. Fig~\ref{fig:Isolation Forest anomaly
  scores over time} shows the anomaly scores for the RSSIs from AP~No.~79
measured at RP~No.~14.
%%% 
\begin{figure}[!tbh]
  \centering%
  \includegraphics[width=\linewidth]{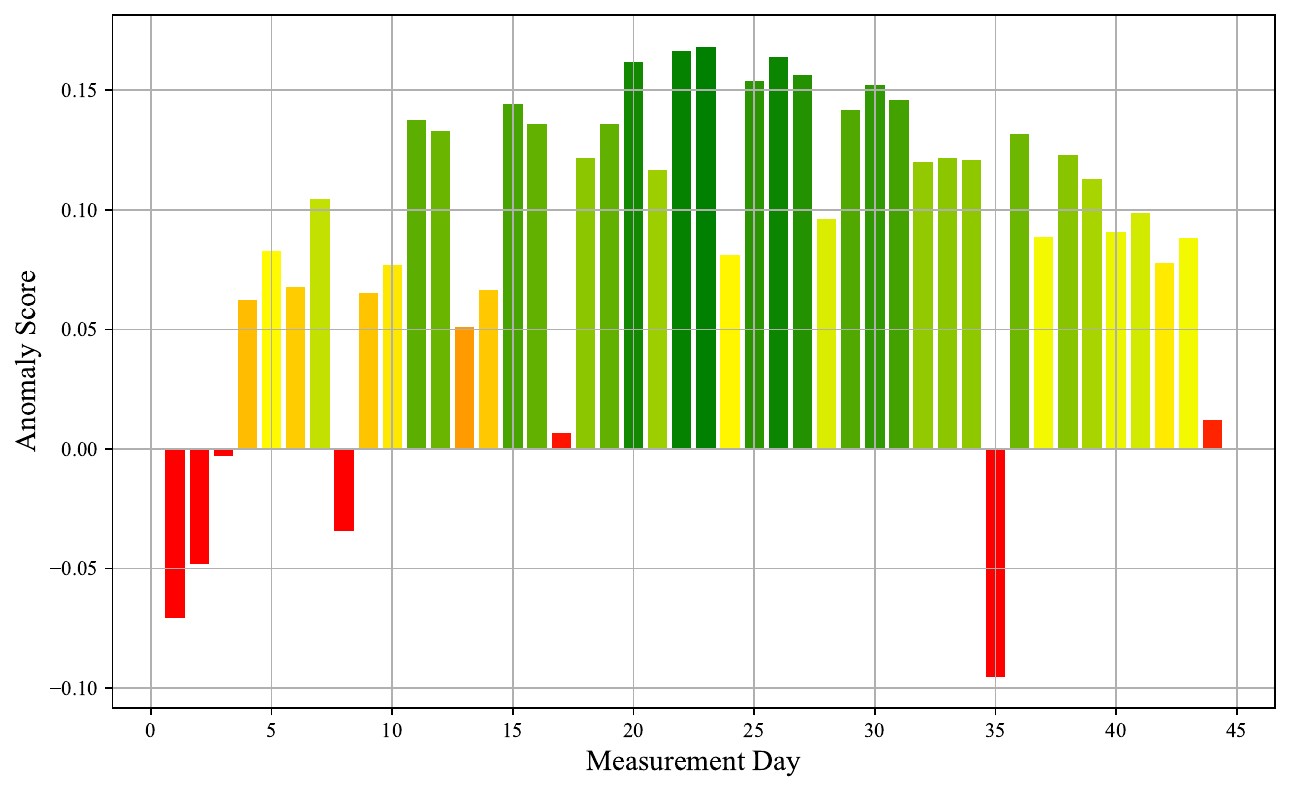}
  \caption{Isolation Forest anomaly scores for the RSSIs from AP~No.~79 measured
    at RP~No.~14 on the 7th floor of the IR building.}
  \label{fig:Isolation Forest anomaly scores over time}
\end{figure}
%%%
The higher, positive anomaly scores (i.e., the green bars) indicate
that the RSSIs on the corresponding measurement days are likely to be normal
given the whole RSSI time series, while the lower, negative anomaly score (i.e.,
the red bars) indicates that the RSSIs on the corresponding measurement
days are likely to be anomalous (e.g., strong interference during the
measurements).

The large fluctuation in RSSIs from a single AP measured at a single RP may not
be directly related with the localization performance. Considering the possible
correlation of the time variations of RSSIs measured at different RPs shown in
Fig.~\ref{fig:RSSI_time}, however, the impact of those fluctuations on the
localization performance can't be simply ignored. Likewise, static databases
based on those anomalous RSSI measurements could negatively affect the actual
performance of indoor localization models trained based on them.

\subsection{Indoor Localization Performance}
\label{sec:indoor-localization-performance}
%%%
We investigate the impact of temporal aspects of Wi-Fi RSSI fingerprints on
indoor localization performance using DNN and Gaussian Process (GP) models,
which are used for the classification of location labels and the regression of location
coordinates, respectively. As we mainly focus on the \textit{changes} of indoor
localization performance over time with dynamic fingerprint databases, we
selected simpler models.

%%%
\begin{table}[!tb]
  \begin{center}
    \caption{Hyperparameters and their values of the DNN model.}
    \label{tbl:dnn_details}
    \begin{threeparttable}
      \begin{tabular}{cc}
        \hline
        Category          & Details                                                                                                                                                                                                                                                                                                                                                                     \\ \hline
        \multicolumn{2}{c}{General Settings}                                                                                                                                                                                                                                                                                                                                                            \\ \hline
        Device            & i9 12900k \& RTX 4090                                                                                                                                                                                                                                                                                                                                                       \\
        Random Seed       & 12345                                                                                                                                                                                                                                                                                                                                                                       \\ \hline
        \multicolumn{2}{c}{Model Settings}                                                                                                                                                                                                                                                                                                                                                              \\ \hline
        Epochs            & SAE\tnote{*}: 30, CLS\tnote{\dag}: 30                                                                                                                                                                                                                                                                                                                                                            \\
        Number of APs     & 465                                                                                                                                                                                                                                                                                                                                                                         \\
        Number of Classes & 30 (RPs)                                                                                                                                                                                                                                                                                                                                                       \\
        Hidden Units      & 512                                                                                                                                                                                                                                                                                                                                                                         \\
        SAE Output Units  & 64                                                                                                                                                                                                                                                                                                                                                                          \\
        Batch Size        & 20                                                                                                                                                                                                                                                                                                                                                                          \\ \hline
        \multicolumn{2}{c}{SAE Model}                                                                                                                                                                                                                                                                                                                                                                   \\ \hline
        Structure         & \begin{tabular}[c]{@{}c@{}}1. Linear layer: 465 -\textgreater 116 (FEA\_DIM/4), \\ Activation: ELU\tnote{\ddag}\\ \\ 2. Linear layer: 116 -\textgreater 64 (SAE\_DIM)\\ \\ 3. Decoder: 64 -\textgreater 116, \\ Activation: ELU\\ \\ 4. Decoder: 116 -\textgreater 465\end{tabular}                                                                                                      \\ \hline
        \multicolumn{2}{c}{CLS Model}                                                                                                                                                                                                                                                                                                                                                                   \\ \hline
        Structure         & \begin{tabular}[c]{@{}c@{}}1. Encoder (from SAE)\\ \\ 2. Linear layer: 64 -\textgreater 512 (HID\_DIM), \\ Activation: ELU\\ \\ 3. Linear layer: 512 -\textgreater 512 (HID\_DIM), \\ Activation: ELU\\ \\ 4. Linear layer: 512 -\textgreater 512 (HID\_DIM), \\ Activation: ELU\\ \\ 5. Output layer: 512 -\textgreater 30 (CLS\_DIM), \\ Batch normalization\end{tabular} \\ \hline
        \multicolumn{2}{c}{SAE Training}                                                                                                                                                                                                                                                                                                                                                                \\ \hline
        Loss Function     & Mean Square Error (MSE)                                                                                                                                                                                                                                                                                                                                                     \\
        Optimizer         & \begin{tabular}[c]{@{}c@{}}Adam with learning \\ rate = \(1 \times 10^{-4}\)\end{tabular}                                                                                                                                                                                                                                                                                   \\ \hline
        \multicolumn{2}{c}{CLS Training}                                                                                                                                                                                                                                                                                                                                                                \\ \hline
        Loss Function     & Cross Entropy Loss                                                                                                                                                                                                                                                                                                                                                          \\
        Optimizer         & \begin{tabular}[c]{@{}c@{}}Adam with learning \\ rate = \(1 \times 10^{-3}\)\end{tabular}                                                                                 
        \\ \hline   
      \end{tabular}
      \begin{tablenotes}
      \item[*] Stacked autoencoder.
      \item[\dag] Classifier.
      \item[\ddag] Exponential linear unit.
      \end{tablenotes}
    \end{threeparttable}
  \end{center}
\end{table}
%%%

\begin{figure}[!tbh]
  \centering%
  \includegraphics[width=\linewidth,trim=0 10 0 10, clip]{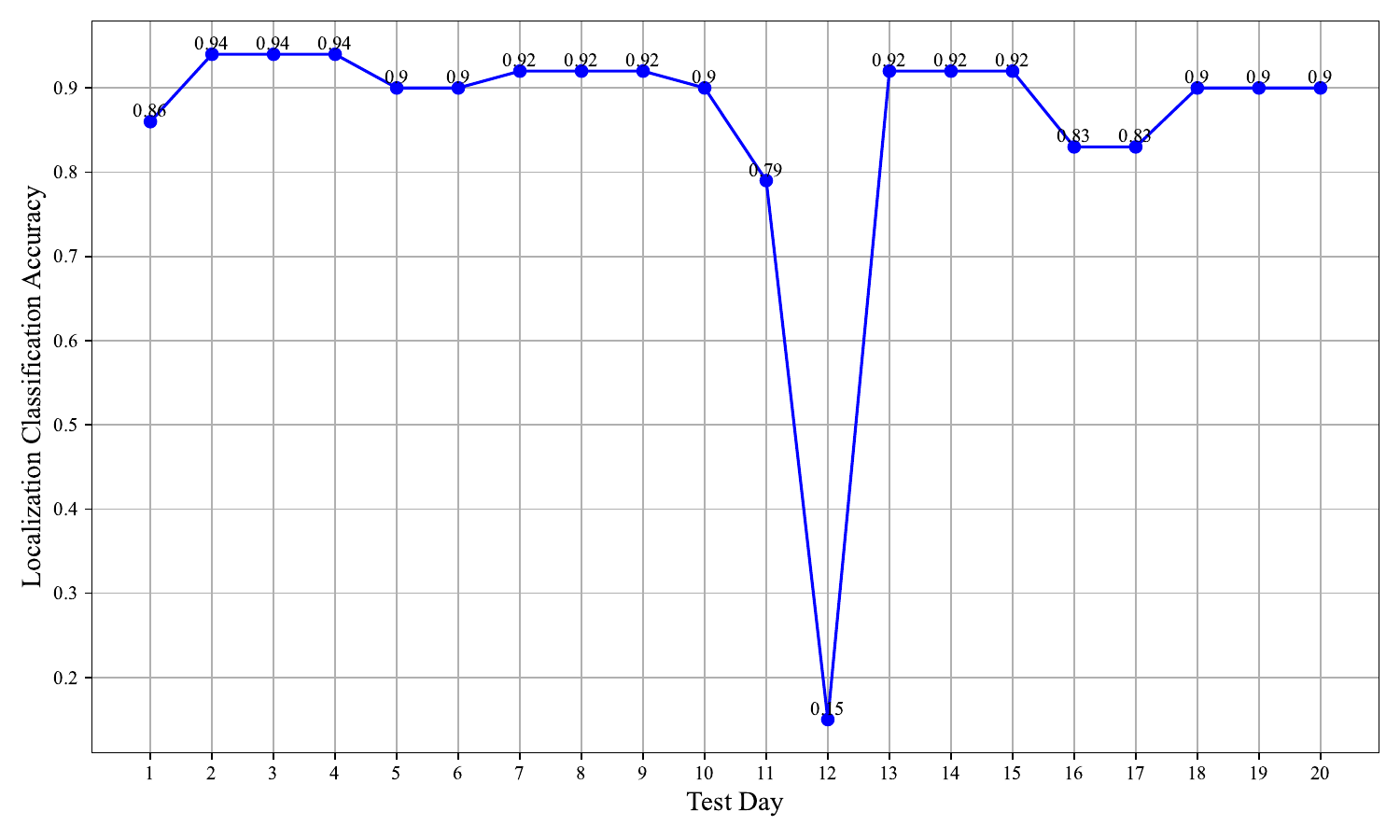}
  \caption{DNN localization accuracy over time.}
  \label{fig:dnn_classification}
\end{figure}

Table~\ref{tbl:dnn_details} summarized the network structure and the
hyperparameter values of the DNN model. With the data measured on the 7th floor
of the IR building in June and July as a training set for a total of 24 days and
those in August as a test set for a total of 20 days, we evaluated the
classification performance of the DNN model trained with the training set
against each daily time slice of the testing set without retraining. The
localization classification accuracy over time is shown in
Fig~\ref{fig:dnn_classification}, where there is a significant drop in the
classification accuracy on the 11th and 12th test days.
%%%
\begin{table}[!tb]
  \begin{center}
    \caption{Hyperparameters and their values of the GP model.}
    \label{tbl:gp_details}
    \begin{tabular}{ccccc}
      \hline
      Category & Details \\ \hline
      Kernel Function & Ornstein-Uhlenbeck (OU) \\
      Variance & 1 \\
      Length scale & 100 \\
      Likelihood & Gaussian \\ \hline
    \end{tabular}
  \end{center}
\end{table}
%%%
%%% 
%%%

\begin{figure}[!tbh]
  \centering%
  \includegraphics[width=\linewidth,trim=0 10 0 10, clip]{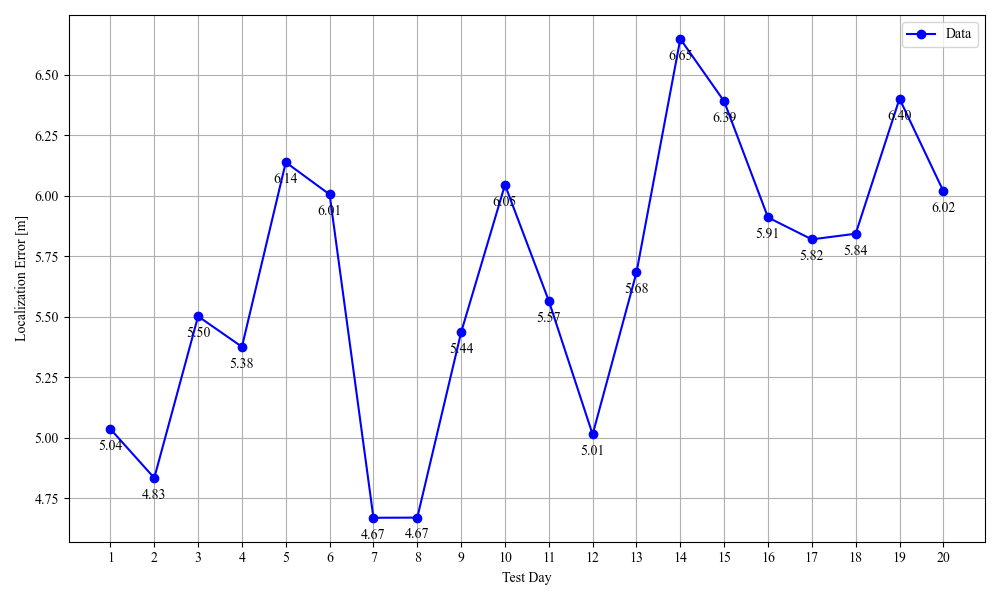}
  \caption{GP localization error over time, where the red dashed line shows the
    results of curve fitting based on a 6th-order polynomial.}
  \label{fig:gp_regression}
\end{figure}

We also implemented a GP model based on GPy~\cite{GPy} and evaluated its
regression performance with the same experimental settings as the DNN model; the
details of the GP model are given in Table~\ref{tbl:gp_details}. The minimum,
the maximum, and the average of the localization errors for a single measurement
day are \SI{4.67}{\m}, \SI{6.65}{\m}, and \SI{5.65}{\m}, respectively. The
regression error over time is shown in Fig~\ref{fig:gp_regression}, where we can
observe a rough cyclic behavior with a long-term trend that the regression error
goes down in the middle of the period of about 5 days and the average over that
period slowly increases over time; Table~\ref{tbl:gp_regression} confirms this
group behavior that the average localization error of a group of 5 days
increases over time, i.e., up to \SI{0.62}{m} from the group of 1--5 days to
that of 16--20 days. These results are consistent with the results of the
analyses based on variance calculations and Isolation Forest in
Section~\ref{sec:isolation-forest} that there are temporal shifts in Wi-Fi
RSSIs.
% In Fig~\ref{fig:gp_regression}, the red dashed line shows
% the error curve fitted using a 6th order polynomial, which is given by
% %%%
% \begin{equation}
%   \label{eq:poly-fitting}
%   \begin{split}
%     \text{Fitted Error} & = 5.561\text{e-}6{\times}x^{6} - 3.188\text{e-4}{\times}x^{5}\\
%                         & + 6.613\text{e-3}{\times}x^{4} - 5.862\text{e-2}{\times}x^{3}\\
%                         & + 1.885\text{e-1}{\times}x^{2} + 5.907\text{e-2}{\times}x\\
%                         & + 4.679.
%   \end{split}
% \end{equation}
% %%%
% The fitted curve extrapolates the localization error to \SI{35.95}{\m} on the
% 25th test day, which suggests that, without retraining, the GP model trained
% with the training set cannot provide reliable location services even for a
% single floor after over 20 days.

%%% 
\begin{table}[!tb]
\centering
\caption{Average localization errors for a group of 5 test days.}
\label{tbl:gp_regression}
\begin{tabular}{cccccc}
\hline
Test Day & All 20 days & 1--5 & 6--10 & 11--15 & 16--20 \\ \hline
Average Error [\si{\m}] & 5.65 &5.37            & 5.36        & 5.86         & 5.99   \\ \hline
\end{tabular}
\end{table}
%%%
%%% 
%%%

As shown in Fig~\ref{fig:gp_regression}, our case study with the XJTLU dynamic
database clearly demonstrates that the localization error increases over time as
the time gap between the measurement times of training and testing datasets
widens, which implies that the model will eventually lose its location service
ability without retraining with the updated training dataset.

\section{Challenges and Opportunities}
\label{sec:challenges-opportunities}
%%%
The discussions in Section~\ref{sec:static-db-issues} and the results from the
case study in Section~\ref{sec:xjtlu-dynamic-db} provide compelling reasons for
the employment of dynamic databases for indoor localization. However, there are
several challenges specific to dynamic databases, which are to be addressed
before their adoption.

One of the major challenges in constructing dynamic databases is to ensure the
access to the same RPs accurately for repetitive RSSI measurements over a long
period of time. Most static databases rely on global coordinates systems like
the universal transverse mercator (UTM) or the world geodetic system 1984
(WGS~84) for specifying RP locations. The advantage of global coordinate systems
is their compatibility and ease of conversion between them. Note that global coordinate 
systems require GPS devices, detailed floor maps of buildings,
and special mapping software. The use of global coordinate systems would be even
more challenging for large-scale indoor localization systems for building
complexes such as hospitals, shopping malls, and transport hubs. In this regard,
the use of a local coordinate system, together with distinct landmarks and
anchor devices, could be a viable option reducing resource overhead and ensuring
the accurate access to the same RPs repeatedly during the construction of
dynamic databases.

Another major challenge is the higher labor costs incurred from repetitive data
measurements over a long period of time. The use of hybrid networks with anchor
devices reduce the labor cost in data measurements because anchor devices can
automatically measure fingerprints and thereby facilitate the measurement cycle
adjustment. Considering that one of the advantages of Wi-Fi fingerprinting is no
requirement for the deployment of new infrastructure or special new user
devices~\cite{Joint}, we recommend to deploy only a limited number of anchor
devices in certain RPs where Wi-Fi signals exhibit substantial fluctuations. As
for anchor devices, we recommend those with a lightweight and straightforward
architecture while capable of the reliable recording of signal variations and
providing sufficient storage space to avoid frequent data transmissions from them.

\section{Concluding Remarks}
\label{sec:concluding-remarks}
%%%
In this paper, we have investigated the implications of the time-varying nature
of Wi-Fi RSSI fingerprints on indoor localization through the case study with
the XJTLU dynamic database. For this case study, we have constructed the XJTLU
dynamic database covering three floors of the IR building on the south campus of
XJTLU, whose Wi-Fi fingerprint data were measured on a daily basis over a period
of 44 days.

The experimental results, with the XJTLU dynamic database, show that the indoor
localization performance of DNN and GP models become worse as the time
difference between the training and the estimation increases; specifically,
there is a noticeable trend of the increase in the localization error of a GP
regression model with the maximum error of 6.65~m after 14 days of training
without model adjustments. In fact, the analyses based on variance calculations
and Isolation Forest indicate the temporal shifts in Wi-Fi RSSIs.
% , and it further increases by \SI{35.95}{\m} after 25 days, which implies that
% the location estimation from the model becomes practically useless after a
% long period of time from the training without any supplementary training.
In this regard, our results of the case study with the XJTLU dynamic database
clearly demonstrate the limitations of static databases and the importance of
the creation and adoption of dynamic databases for future indoor localization
research and real-world deployments.

Based on our hands-on experience obtained during the construction of the XJTLU
dynamic database, we have also proposed guidelines for the cost-effective
construction of dynamic databases, where we encourage the adoption of local
coordinates and the integration of hybrid network data, the latter of which
enables researchers to calibrate their algorithms leveraging dynamic databases
and to handle issues resulting from the differences between simple laboratory
and more complicated deployment environments.

\section*{Acknowledgment}
This work was supported in part by the Postgraduate Research Scholarship (under
Grant PGRS1912001), the Key Program Special Fund (under Grant KSF-E-25), and the
Research Enhancement Fund (under Grant REF-19-01-03) of Xi'an Jiaotong-Liverpool
University.

%%%
\balance % to balance the columns
%%%

% %%% References
% \bibliographystyle{IEEEtran}%
% \bibliography{IEEEabrv,reference}%

\end{document}